%% file: paper.tex
\documentclass[dvipsnames]{article}
\usepackage{template_conference}

\usepackage{booktabs}
\usepackage{graphicx}
\usepackage{enumitem}
\usepackage{wrapfig}
\usepackage{needspace}
\usepackage{placeins}
\usepackage{algorithm}
\usepackage{algpseudocode}
\usepackage[utf8]{inputenc}
\usepackage[T1]{fontenc}
\DeclareUnicodeCharacter{FF5E}{\textasciitilde}

\usepackage{adjustbox}
\usepackage{amsmath}
\usepackage{colortbl}
\definecolor{lightgray}{rgb}{0.9,0.9,0.9}
\usepackage{caption}
\usepackage{subcaption}
\usepackage[svgnames]{xcolor}
\usepackage{setspace}
\usepackage{url}
\usepackage{multirow}
\usepackage{tabularx}
\usepackage{blindtext}
\usepackage{pgfplots}
\pgfplotsset{compat=1.18} 
\usepackage{tikz}
\usetikzlibrary{er,positioning,arrows.meta,calc}
\usepackage{makecell}
\usepackage{tipa}
\tipasafemode
\usepackage{siunitx}
\usepackage{tocloft}
\usepackage{listings}
\usepackage{xurl}
\usepackage{rotating}
\usepackage[normalem]{ulem}
\usepackage{multicol}
\usepackage{titlesec}
\usepackage{tcolorbox}
\tcbuselibrary{breakable,theorems}
\usepackage{transparent}
\usepackage{mathtools}

\usepackage{amsthm}
\usepackage{amsfonts}
\usepackage{etoc}
\usepackage{color}
\usepackage{autobreak}
\usepackage{soul}
\usepackage{amssymb}
\usepackage{bbding}

\newcommand{\method}{UMM-JEPA}
\newcommand{\methodcore}{UMM-JEPA-core}
\newcommand{\gain}[1]{{\scriptsize\textcolor{blue}{(#1)}}}

\newtheorem{proposition}{Proposition}
\newtheorem{lemma}{Lemma}

\newtcolorbox{codeblock}{
  colback=gray!10,   
  colframe=gray!50,  
  boxrule=0.5mm,     
  arc=2mm,           
  left=5pt,          
  top=5pt,          
  bottom=5pt,        
}

\newtcbtheorem{definition}{Definition}{
  enhanced,
  colback=cyan!5!white,
  colframe=cyan!45!blue!60,
  fonttitle=\bfseries,
  fontupper=\normalfont,
  arc=4mm,
  boxrule=1pt,
  drop shadow southwest,
  attach boxed title to top left={xshift=6mm, yshift=-2mm},
  boxed title style={
    colback=cyan,
    size=small,
    arc=2mm,
    boxrule=0.8pt,
    left=2mm, right=2mm
  },
  before skip=10pt,
  after skip=10pt,
  left=5mm,
  right=5mm,
  top=3mm,
  bottom=3mm
}{cor}

\useunder{\uline}{\ul}{}

\input{math_commands.tex}

\newcommand*\justify{
  \fontdimen2\font=0.4em
  \fontdimen3\font=0.2em
  \fontdimen4\font=0.1em
  \fontdimen7\font=0.1em
  \hyphenchar\font=`\-
}

\renewcommand{\texttt}[1]{
  \begingroup
  \ttfamily
  \begingroup\lccode`~=`/\lowercase{\endgroup\def~}{/\discretionary{}{}{}}
  \begingroup\lccode`~=`[\lowercase{\endgroup\def~}{[\discretionary{}{}{}}
  \begingroup\lccode`~=`.\lowercase{\endgroup\def~}{.\discretionary{}{}{}}
  \catcode`/=\active\catcode`[=\active\catcode`.=\active
  \justify\scantokens{#1\noexpand}
  \endgroup
}

\definecolor{blueviolet}{RGB}{138,43,226}
\newtcolorbox{insightblock}{
  colback=blueviolet!5,
  colframe=blueviolet!50!black!50!,
  boxrule=0.5mm,
  arc=2mm,
  left=0pt,
  right=8pt,
  top=8pt,
  bottom=8pt,
}

\newtcolorbox{abstractbox}{
    colback=blue!5!white,
    frame empty,
    boxrule=1pt,
    arc=4mm,
    left=8pt,
    right=8pt,
    top=8pt,
    bottom=8pt,
    opacityback=0.9
}

\newtcolorbox{examplebox}[1][]{
    breakable,
    colback=black!4,
    colframe=black!35,
    boxrule=0pt,
    sharp corners,
    boxsep=0pt,
    left=8pt,
    right=7pt,
    top=6pt,
    bottom=6pt,
    before upper={
        \if\relax\detokenize{#1}\relax
        \else
            {\small\bfseries\color{black!85} #1\par}
            \vspace{3pt}
        \fi
    },
    before skip=0.8\baselineskip,
    after skip=0.8\baselineskip,
    pad at break*=1.2mm
}

\title{UMM-JEPA: Generative Pretraining via Masked Next-Scale Representation Prediction for Unified Multimodal Models}

\author{
\footnotesize
\textbf{Yucheng Zhou}\textsuperscript{1} \quad
\textbf{Yunjia Xi}\textsuperscript{2} \quad
\textbf{Di Yin}\textsuperscript{2} \quad
\textbf{Jianbing Shen}\textsuperscript{1}
\\[2pt]
\scriptsize \textsuperscript{1}University of Macau \quad
\textsuperscript{2}Tencent Youtu Lab
\\[2pt]
\scriptsize \texttt{yucheng.zhou@connect.um.edu.mo}
}

\begin{document}

\maketitle

\bibliographystyle{template_conference}
\input{main}

\end{document}

%% file: math_commands.tex
\usepackage{amsmath,amsfonts,bm}

\def\eqref#1{equation~\ref{#1}}

\def\1{\bm{1}}

\DeclareMathAlphabet{\mathsfit}{\encodingdefault}{\sfdefault}{m}{sl}
\SetMathAlphabet{\mathsfit}{bold}{\encodingdefault}{\sfdefault}{bx}{n}



%% file: main.tex
\begin{abstractbox}
\begin{center}
\textbf{\Large Abstract}
\end{center}
Unified multimodal models can support both visual understanding and generation, yet generative training does not necessarily improve understanding. We argue that this limitation may arise from a mismatch between the representation targeted by generation and the perceptual representation used for understanding. Conventional generation objectives primarily optimize synthesis-oriented spaces such as discrete visual codes or generative latents. We therefore introduce \method{}, which reformulates generation as progressive prediction of continuous visual-understanding representations. The model constructs a multi-scale representation hierarchy from coarse to fine, while pixel rendering is decoupled from representation pretraining and attached only afterward.
Controlled comparisons show that the representation-prediction objective accounts for most of the understanding improvement, while a data-matched control leaves the group means nearly unchanged. \method{} reaches a 16-benchmark visual mean of 51.71, against 46.36 for the understanding-only Qwen-VL baseline, and the text mean also rises. Compared with discrete-code and latent-flow generation objectives, perceptual representation prediction yields stronger and more consistent transfer to understanding. Linear probes and visual-evidence interventions further reveal richer object and spatial structure and stronger use of image-specific information. These results suggest that generation can benefit understanding when it directly supervises the perceptual representations that understanding consumes.
\end{abstractbox}
\vspace{4pt}

\section{Introduction}
\label{sec:intro}

Unified multimodal large language models (MLLMs) increasingly support both visual understanding and generation within a single system \citep{chameleon,emu3,bagel,onecat,showo2,chen2026unify,zhou2026multimodal}. Coexistence, however, does not imply positive knowledge transfer: a model may learn to understand and generate images without generative training actually improving the representations used for understanding. Recent controlled studies reveal a notable asymmetry in this knowledge flow: visual-understanding supervision can substantially benefit visual generation, whereas simply increasing visual-generation training does not consistently improve visual understanding \citep{yang2025self,han2026towards,zhou2026world}. This raises the central question of this paper: \emph{how should generation be formulated if it is expected to improve understanding?}

\begin{figure}[t]
\centering
\includegraphics[width=\linewidth]{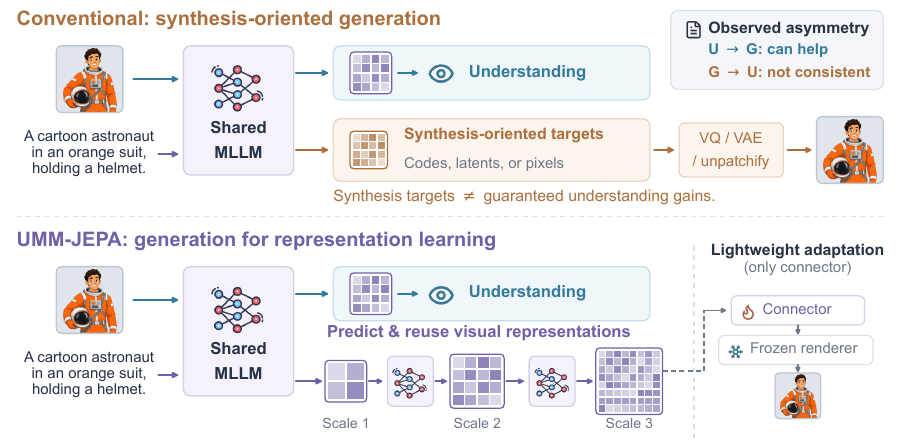}
\caption{\small \textbf{Reformulating generation as representation learning.}
Conventional unified models optimize synthesis-oriented targets, whereas \method{} directly predicts continuous visual-understanding representations through progressive coarse-to-fine construction. Pixel rendering is decoupled from core representation pretraining.}
\label{fig:motivation}
\end{figure}

We hypothesize that this asymmetry originates from a mismatch between the representation targeted by generation and the representation consumed by understanding. Conventional unified models typically formulate generation in synthesis-oriented spaces, such as discrete visual codes, generative VAE latents, or pixel-oriented objectives. Visual understanding, in contrast, relies on perceptual representations, semantically organized visual states produced by a vision encoder and consumed by the language model. Under conventional generation objectives, generative supervision may update shared model parameters without directly shaping the perceptual state that understanding later reads. The key issue may therefore be not whether generation is introduced, but \emph{what representation the generative objective forces the model to construct}.

This perspective motivates us to reformulate generation as \emph{progressive visual representation construction}. Instead of treating perceptual features as intermediate conditioning for image synthesis, we make prediction of those perceptual representations the generation objective itself.

We instantiate this formulation as \method{}, a unified multimodal model whose understanding and representation-generation paths share a Qwen3 LLM and a SigLIP2 visual front end \citep{qwen3,siglip2}. Instead of predicting pixels, discrete visual codes, or dedicated generative latents, \method{} progressively predicts continuous SigLIP2 representations across four spatial scales ($64^2$ to $512^2$ resolution, corresponding to $4$ to $256$ visual tokens). During training, the coarse-to-fine conditional mappings are learned under teacher-forced visual context; during inference, predicted coarse states are recursively reused through the shared backbone to construct finer states. The formulation follows the predictive-representation principle of JEPA \citep{ijepa}, but extends it to conditional multimodal generation. Pixel rendering is deliberately removed from core representation pretraining and attached only afterward through a lightweight connector and a pretrained diffusion renderer.

We evaluate this hypothesis through a controlled evidence chain. An understanding-only Qwen-VL baseline establishes whether generative representation learning provides any benefit, while a data-matched control separates the effect of the objective from that of additional image--text data. Generation-paradigm controls further compare continuous perceptual prediction against discrete visual-code prediction (UMM-VQ) and latent-flow generation (Show-o2), and component ablations examine bidirectional visual attention, masking, and auxiliary reconstruction. Finally, linear probes and visual-evidence interventions analyze how generative representation learning changes the learned visual states and their use during multimodal understanding.
Our contributions are:
\begin{itemize}[leftmargin=*, topsep=0pt, itemsep=0pt]

\item We identify the representation targeted by generation as a key factor governing whether generative pretraining transfers to visual understanding, highlighting the importance of supervision that directly shapes the perceptual representation space used by understanding.

\item We introduce \method{}, which reformulates visual generation as progressive next-scale prediction of continuous visual-understanding representations while decoupling pixel rendering from core representation pretraining.

\item We provide controlled empirical evidence separating the effects of additional data, attention design, and generation paradigm, showing that perceptual representation prediction transfers more effectively to understanding than the compared synthesis-oriented objectives.

\item We provide mechanistic evidence that generative representation learning changes both the structure of visual states and how visual information is used during reasoning, through linear probes and visual-evidence interventions.

\end{itemize}

\section{Related work}
\label{sec:related}

Unified multimodal models adopt markedly different generative parameterizations, including autoregressive discrete visual tokens \citep{chameleon,emu3,janus,song2026broad,zhou2026condition}, next-scale visual autoregression \citep{var}, latent flow matching and diffusion-based synthesis \citep{showo2,chen2025dcae,chen2025towards,yang2025dc,yang2026hicogen,wang2026ladr}, and modality-specialized, mixture-of-transformer, or partially separated pathways \citep{mot,onecat,bagel,chen2026unify,zhou2026multimodal}. Controlled studies report an asymmetric knowledge flow, understanding supervision benefits generation, whereas extra generation training does not consistently improve understanding and gains depend on the task and reasoning regime \citep{yang2025self,wang2025complexbench,han2026towards,wen2026unig2u,wu2026synergy,zhou2026world}, and auxiliary generative targets such as depth or segmentation can benefit perception \citep{su2026unimrg}. Closest to us, semantically meaningful representations have been placed inside the generative pathway via continuous visual embeddings \citep{emu,emu2}, diffusion in a CLIP feature space \citep{blip3o}, or representation prediction as intermediate generative states \citep{representationforcing}, under the principle of predictive learning in representation space \citep{ijepa}. We instead make prediction of the visual-understanding representation itself the pretraining objective and remove pixel rendering from core pretraining, isolating whether transfer strengthens when generative supervision shapes the representation that understanding consumes; Appendix~\ref{app:related} gives a detailed discussion.

\section{Method}
\label{sec:method}

\method{} reformulates visual generation as progressive prediction of the continuous perceptual representations used by multimodal understanding. Understanding and representation generation share the same multimodal backbone, while a frozen visual encoder provides the prediction targets.

\begin{figure}[t]
    \centering
    \includegraphics[width=\linewidth]{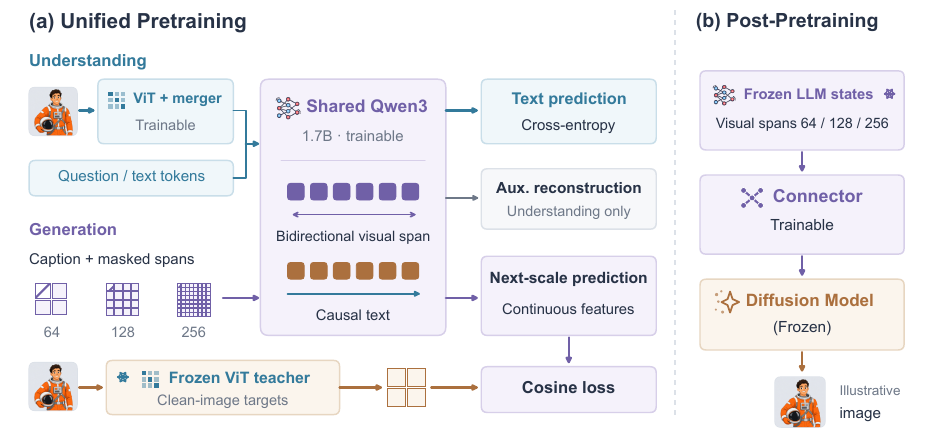}
    \caption{\small \textbf{Overview of \method{}.}
Understanding and representation generation share a single multimodal backbone. Understanding uses language-modeling supervision, while generation progressively predicts frozen perceptual representations from coarse to fine. Pixel rendering is decoupled from representation pretraining and attached only afterward through a separately trained connector and pretrained diffusion renderer.}
    \label{fig:method}
\end{figure}

During pretraining, understanding samples use language-modeling supervision, whereas generation samples learn masked coarse-to-fine representation prediction. Pixel rendering is deliberately excluded from representation pretraining and attached only afterward through a lightweight connector and a pretrained diffusion renderer. Figure~\ref{fig:method} summarizes the framework.

\subsection{Design Motivation for Representation Prediction}
\label{sec:method:theory}

\noindent\textbf{Perceptual representation as the prediction target.}
Let $\mathbf{z}=E(\mathbf{x})$ denote the perceptual representation produced by the visual encoder and consumed by the multimodal understanding model $h$. Our central question is which generation target should be optimized if generative supervision is expected to improve the representation used by understanding.

\begin{proposition}[Target-aligned representation prediction]
\label{prop:target}
Assume that the downstream understanding mapping $h$ is locally $L$-Lipschitz in the perceptual representation space. Then
\begin{align}
    \|h(\hat{\mathbf{z}})-h(\mathbf{z})\|
    \le
    L\|\hat{\mathbf{z}}-\mathbf{z}\|.
\end{align}
Thus, prediction error measured directly in the representation consumed by understanding provides direct control over the induced downstream deviation.
\end{proposition}

\noindent\textbf{Design implication.}
We therefore use frozen perceptual representations from the visual encoder as the generation target, rather than supervising generation only through a separate synthesis-oriented space.

\begin{lemma}[Conditioning on coarse states cannot increase optimal prediction risk]
\label{lem:coarse}
Let $\mathbf{Z}^{k+1}$ denote a finer-scale representation and $\mathcal{C}_k=(C,\mathbf{Z}^{64},\ldots,\mathbf{Z}^{k})$ the text condition together with available coarser states. Under squared prediction loss,
\begin{align}
    \mathcal{R}^*(\mathbf{Z}^{k+1}\mid\mathcal{C}_k)
    \le
    \mathcal{R}^*(\mathbf{Z}^{k+1}\mid C).
\end{align}
\end{lemma}

\noindent\textbf{Design implication.}
We therefore factorize representation generation progressively, conditioning each finer scale on the previously available coarse states.

\begin{proposition}[Masked prediction induces conditional recovery]
\label{prop:mask}
Let $M$ denote masked visual states and $\bar M$ the visible states. Under squared prediction loss, the Bayes-optimal predictor is
\begin{align}
    f^*(C,\mathbf{Z}_{\bar M})
    =
    \mathbb{E}[\mathbf{Z}_M\mid C,\mathbf{Z}_{\bar M}].
\end{align}
Masking therefore removes direct access to the target and requires recovery from textual, spatial, and cross-scale context.
\end{proposition}

\noindent\textbf{Design implication.}
We randomly mask visual states during training to prevent trivial direct observation of prediction targets. We additionally include fully masked samples so that the text-to-visual transition required at the start of inference is explicitly represented during training.

\noindent\textbf{Relation to the implemented objective.}
The analysis above uses squared loss for clarity. Because the target representations are $\ell_2$-normalized, squared Euclidean distance is equivalent to cosine distance up to a constant. We therefore optimize cosine representation loss in practice. Formal proofs and assumptions are provided in Appendix~\ref{app:theory}.

\subsection{Progressive Masked Representation Prediction}
\label{sec:method:prediction}

\noindent\textbf{Representation targets.}
Let $\mathbf{z}^{k}$ denote the frozen SigLIP2 (layer 23) representation of the ground-truth image at resolution $k\times k$, where $k\in\{64,128,256,512\}$, corresponding to $4/16/64/256$ visual tokens. The teacher encoder is frozen and applied independently to clean target images.

\noindent\textbf{Progressive factorization.}
Generation samples follow the coarse-to-fine chain
\begin{align}
p(\mathbf{z}^{64} \mid c),\qquad
p(\mathbf{z}^{128} \mid c, \mathbf{z}^{64}),\qquad
p(\mathbf{z}^{256} \mid c, \mathbf{z}^{64}, \mathbf{z}^{128}),\qquad
p(\mathbf{z}^{512} \mid c, \mathbf{z}^{64}, \mathbf{z}^{128}, \mathbf{z}^{256}),
\end{align}
where $c$ is the conditioning context (the language caption and any visible visual states under masking). Each parent predicts its four native children at the next scale, supervised by a cosine loss against the frozen teacher, with the four scales equally weighted and normalized.

\noindent\textbf{Training: teacher-forced context.}
Training is teacher-forced. The 64/128/256 input spans are constructed from the ground-truth image through the student visual pathway (encoder, merger, and masking recipe); they are never obtained by feeding back the model's preceding predictions. The shared backbone contextualizes these inputs under cross-scale causal attention, with bidirectional attention within each visual span, and a prediction head maps each parent hidden state to the features of its four spatially corresponding children, supervised by clean-image features from the separate frozen teacher. The 4-to-16-token transition is therefore learned conditional prediction through the backbone and head, not interpolation or duplication of a coarse grid. The $512$-scale enters the sequence only as a prediction target.

\noindent\textbf{Inference: recursive construction.}
At inference no ground-truth visual context is available. Starting from the caption, predicted states $\hat{\mathbf{z}}^{64}$, $\hat{\mathbf{z}}^{128}$, and $\hat{\mathbf{z}}^{256}$ are recursively reused as context for subsequent prediction through the same shared backbone,
\begin{align}
\hat{\mathbf{z}}^{64}\rightarrow\hat{\mathbf{z}}^{128}\rightarrow\hat{\mathbf{z}}^{256}\rightarrow\hat{\mathbf{z}}^{512},
\end{align}
so that a predicted visual state is both a generative output and a subsequent perceptual input. Conditioning each scale on its predecessors also distinguishes the objective from predicting all scales independently from the caption. The resulting train--test distinction is therefore explicit: training learns the conditional mappings using teacher-forced visual context, whereas inference composes the same mappings recursively using model-predicted states.
\\
\noindent\textbf{Masking.}
During training, visual states are randomly masked to expose the model to both missing-context recovery and coarse-to-fine refinement. In addition to partial masking, we include fully masked samples to train the text-only initialization required at the first generation scale, and unmasked samples to explicitly train refinement from available coarse states. Both masked and visible parents receive representation supervision. Detailed masking ratios and loss weights are provided in Appendix~\ref{app:impl}. Figure~\ref{fig:nsp} illustrates the resulting training and inference procedures.

\begin{figure}[t]
    \centering
    \includegraphics[width=\linewidth]{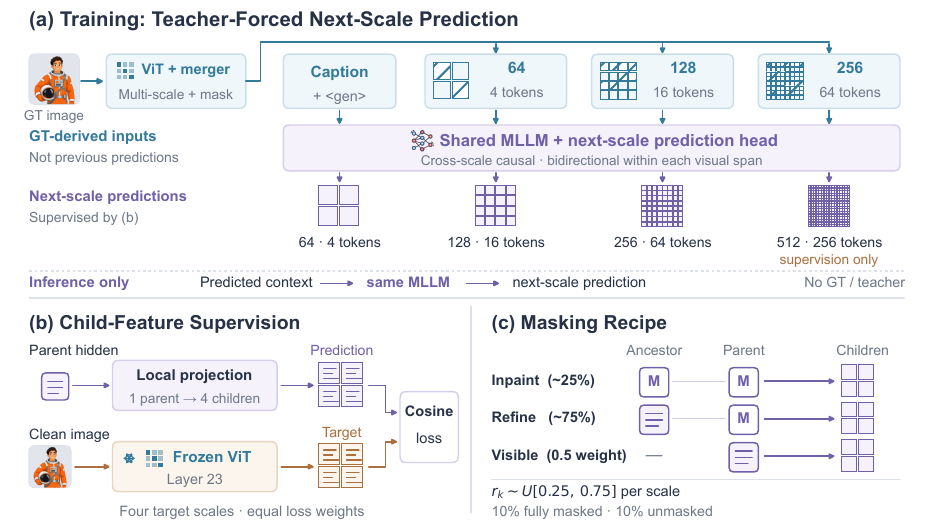}
    \caption{\small \textbf{Progressive masked representation prediction.}
Each parent predicts four child representations at the next scale under
frozen-teacher supervision, conditioned on ground-truth visual states during
training and on recursively reused predictions at inference. Masking trains
missing-context recovery and coarse-to-fine refinement.}
    \label{fig:nsp}
\end{figure}

\subsection{Joint Multimodal Pretraining}
\label{sec:method:joint}
Understanding and representation generation are optimized jointly within one backbone over a mixture of understanding and generation samples,
\begin{align}
    \mathcal{L}
    =
    \mathcal{L}_{\mathrm{LM}}
    +
    \lambda_{\mathrm{rep}}\,\mathcal{L}_{\mathrm{rep}}
    +
    \lambda_{\mathrm{rec}}\,\mathcal{L}_{\mathrm{rec}}.
\end{align}
\noindent\textbf{Understanding supervision.}
Understanding samples use standard autoregressive language-modeling supervision: the model predicts response tokens from the preceding text and visual tokens.
\\
\noindent\textbf{Representation generation.}
Generation samples are optimized using the masked progressive prediction objective of Section~\ref{sec:method:prediction}. All multimodal backbone parameters are shared with the understanding pathway.
\\
\noindent\textbf{Auxiliary reconstruction.}
For understanding samples, \method{} additionally appends a text-conditioned reconstruction tail that predicts the sample's multi-scale SigLIP2 representations without access to its original visual tokens; the reconstruction uses the same progressive representation objective and acts only as an auxiliary understanding-side signal. This component is not required for the central generation-to-understanding effect: we therefore use \methodcore{}, which removes reconstruction, for causal attribution of the representation-prediction objective, and reserve \method{} for final performance reporting. Its contribution is separately quantified in Section~\ref{sec:exp:attribution}, and implementation details are provided in Appendix~\ref{app:impl:rec}.
\\
\noindent\textbf{Shared attention structure.}
Visual tokens attend bidirectionally within each visual span, while text tokens remain causal \citep{pytorch_flexattention}, allowing each visual span to form a complete internal representation before being read by text.

\subsection{Decoupled Image Rendering}
\label{sec:method:rendering}

\noindent\textbf{Rendering is post-pretraining.}
Pixel rendering does not participate in representation pretraining. After the multimodal representation backbone is learned, we freeze it and attach a lightweight connector together with a pretrained FLUX.2 renderer \citep{flux2}; the connector maps the learned multimodal visual states to the conditioning space of the renderer. Representation pretraining and rendering are therefore two separate procedures with different training mechanisms: during representation pretraining (Sections~\ref{sec:method:prediction} and~\ref{sec:method:joint}) the visual context is teacher-forced and the supervision is applied in representation space, whereas during rendering the visual context is produced by the model itself and the supervision is applied in pixel space. This separation ensures that the pixel-space objective does not alter the representation-learning mechanism studied in this paper.

\noindent\textbf{Connector alignment with a frozen renderer.}
In the first rendering stage, the representation backbone and the renderer are frozen and only the connector is optimized, on images sampled from the same image--text pool used for representation pretraining. The connector is fitted through the renderer's native rectified-flow objective. Training gradually transitions from teacher-forced visual conditioning toward prediction-feedback conditioning: the feedback probability is zero during early training and is then ramped up over a short interval, so that the connector is eventually fitted to the model-predicted visual states that are actually used at inference.

\noindent\textbf{Prediction-feedback renderer adaptation.}
The representation backbone remains frozen. We jointly optimize the connector and lightweight LoRA adapters \citep{hu2021lora} inserted into the text-conditioning modules of the diffusion transformer \citep{chen2025towards,yang2025dc}, while the remaining renderer parameters stay frozen. To match recursive generation at inference and mitigate condition errors between teacher-forced training and recursive rollouts \citep{zhou2026condition,yang2026hicogen}, each training step uses a two-pass procedure. In the first pass, the frozen representation pathway predicts the multi-scale visual conditions without gradient updates, and the resulting one-step estimates are spliced back into the conditioning sequence as visual context. The second pass recomputes the forward pass from this model-predicted context and evaluates the diffusion velocity objective, and only this second-pass loss is back-propagated into the connector and the LoRA adapters. The rendering pathway is thus optimized under model-generated rather than purely teacher-forced visual conditions. A final generation instruction-tuning stage continues this training while additionally adapting low-rank adapters on the language backbone's attention projections. Text-only Qwen, Qwen-VL, and \method{} all use this recipe; their scores in Section~\ref{sec:exp:generation} are taken after this stage. Numerical configurations of all rendering stages are given in Appendix~\ref{app:impl:render}.

\section{Experiments}
\label{sec:exp}

\subsection{Experimental setup}
\label{sec:exp:setup}

We train every model in Table~\ref{tab:main} from the same pretrained Qwen3-1.7B checkpoint \citep{qwen3}, on 10B text and 45B multimodal tokens. All of them also use SigLIP2 \citep{siglip2}, except Qwen-EF, which keeps that LLM initialization but replaces the pretrained ViT with a patch embedding trained from scratch. Generation-bearing models add 10B image--text tokens, isolated in Section~\ref{sec:exp:attribution}. Qwen-VL is understanding-only. Qwen-EF is an architecture reference, not a comparison of generation objectives. UMM-VQ and Show-o2 retrain discrete-code and latent-flow generation under the same extra budget. \methodcore{} is masked next-scale representation prediction with bidirectional visual attention; \method{} adds auxiliary reconstruction. \emph{Text}, \emph{Visual Understanding}, and \emph{Visual Specialty} are means of 4, 7, and 9 benchmarks. GenEval and DPG are kept separate. Additional experimental and evaluation details are in Appendices~\ref{app:setup} and~\ref{app:impl}.

\subsection{Does representation-based generative pretraining improve understanding?}
\label{sec:exp:main}

\method{} reaches a 16-benchmark visual mean of 51.71 and a text mean of 61.29, compared with 46.36 and 59.16 for Qwen-VL (Table~\ref{tab:main}). Qwen-EF is substantially weaker on visual benchmarks, so the pretrained encoder is held fixed in the comparisons below.

\noindent\textbf{Where the gains concentrate.}
The largest gains are on visually grounded specialty tasks: ChartQA ($+15.44$), RefCOCO ($+13.91$), DocVQA ($+11.73$), and VSR ($+6.54$), with OCRBench and TextVQA following the same trend. The specialty mean rises from 45.69 to 53.84, while visual understanding rises more moderately, from 47.23 to 48.97.

\noindent\textbf{No language cost.}
Text improves as well, from 59.16 to 61.29, led by MATH-500 ($+5.40$) and MGSM-zh ($+2.40$). Under this protocol, generative representation pretraining improves text and visual understanding together.

\begin{table}[t]\small
\centering
\small
\caption{\small Main pretraining comparison. Every system, including Qwen-EF, is trained by us from the same pretrained Qwen3-1.7B checkpoint. Qwen-EF keeps that initialization and replaces only the ViT with a patch embedding trained from scratch. The other systems also share SigLIP2; generation-bearing models add 10B image--text data (Table~\ref{tab:attribution}). Bold is best among the five systems. Blue parentheses are the gain of \method{} over Qwen-VL. HallusionBench uses aAcc.}
\label{tab:main}
\vspace{-2mm}
\begin{tabular}{llccccc}
\toprule
 & Benchmark & Qwen-EF & Qwen-VL & UMM-VQ & Show-o2 & \method{} \\
\midrule
\multirow{4}{*}{\shortstack[l]{Text\\understanding}}
 & MMLU                & 49.54 & 51.23 & 50.95 & 51.27 & \textbf{51.57} \gain{+0.34} \\
 & GSM8K               & 76.12 & 76.42 & 74.60 & 74.60 & \textbf{76.80} \gain{+0.38} \\
 & MGSM-zh             & 59.20 & 59.20 & 60.80 & 60.40 & \textbf{61.60} \gain{+2.40} \\
 & MATH-500            & 53.20 & 49.80 & 52.20 & 49.60 & \textbf{55.20} \gain{+5.40} \\
\midrule
\multirow{7}{*}{\shortstack[l]{Visual\\understanding}}
 & MMMU                & 43.72 & 41.61 & 44.67 & 44.11 & \textbf{45.11} \gain{+3.50} \\
 & MMMU-Pro            & 20.17 & 20.64 & 20.95 & 21.24 & \textbf{21.50} \gain{+0.86} \\
 & MMBench EN          & 14.86 & \textbf{47.76} & 47.14 & 47.69 & \textbf{47.76} \gain{+0.00} \\
 & MMBench CN          & 12.00 & 33.98 & 37.27 & 37.30 & \textbf{37.54} \gain{+3.56} \\
 & SEED-Bench          & 40.13 & 60.37 & 61.64 & 62.13 & \textbf{62.42} \gain{+2.05} \\
 & SEED-Bench-2+       & 42.69 & 51.43 & 52.48 & 52.00 & \textbf{52.88} \gain{+1.45} \\
 & ScienceQA           & 69.21 & 74.81 & 75.09 & 75.06 & \textbf{75.61} \gain{+0.80} \\
\midrule
\multirow{9}{*}{\shortstack[l]{Visual\\specialty}}
 & ChartQA             & 9.32 & 41.20 & 56.36 & 56.40 & \textbf{56.64} \gain{+15.44} \\
 & OCRBench            & 2.70 & 48.30 & 53.10 & 52.50 & \textbf{55.90} \gain{+7.60} \\
 & TextVQA             & 5.78 & 48.83 & 57.70 & 57.89 & \textbf{58.12} \gain{+9.29} \\
 & DocVQA              & 6.10 & 31.36 & 43.06 & 43.02 & \textbf{43.09} \gain{+11.73} \\
 & HallusionBench      & 54.26 & 60.25 & 60.78 & 61.72 & \textbf{62.04} \gain{+1.79} \\
 & VSR                 & 51.88 & 49.02 & 52.78 & 52.70 & \textbf{55.56} \gain{+6.54} \\
 & ViewSpatial         & 28.17 & 31.48 & 32.62 & 33.07 & \textbf{34.21} \gain{+2.73} \\
 & RefCOCO             & 7.55 & 32.62 & 37.80 & 38.10 & \textbf{46.53} \gain{+13.91} \\
 & CountBench          & 11.50 & 68.17 & 72.28 & 71.87 & \textbf{72.48} \gain{+4.31} \\
\bottomrule
\end{tabular}
\vspace{-2mm}
\end{table}

\begin{table}[t]\small
\centering
\caption{\small Attribution and component analysis. ``Extra 10B'' is image--text data beyond the shared corpus. NSP arrives together with this data, so it is read against the data-matched row. The no-mask variant is scored on understanding only. Per-benchmark results are in Appendix~\ref{app:fullablation}.}
\label{tab:attribution}
\vspace{-2mm}
\begin{tabular}{lcccccccc}
\toprule
Variant & Extra 10B & Bi-Attn & NSP & Mask & Recon & Text $\uparrow$ & VisU $\uparrow$ & VisS $\uparrow$ \\
\midrule
Qwen-VL                   & $\times$     & $\times$     & $\times$     & --          & $\times$     & 59.16 & 47.23 & 45.69 \\
Qwen-VL data-matched      & $\checkmark$ & $\times$     & $\times$     & --          & $\times$     & 58.90 & 47.27 & 45.65 \\
Qwen-VL + Bi-Attn         & $\times$     & $\checkmark$ & $\times$     & --          & $\times$     & 59.65 & 47.79 & 45.81 \\
w/o Mask                  & $\checkmark$ & $\checkmark$ & $\checkmark$ & $\times$    & $\times$     & 59.90 & 48.42 & 51.43 \\
\methodcore{} (w/o Recon) & $\checkmark$ & $\checkmark$ & $\checkmark$ & $\checkmark$ & $\times$   & 59.89 & 48.55 & 52.62 \\
\method{}                 & $\checkmark$ & $\checkmark$ & $\checkmark$ & $\checkmark$ & $\checkmark$ & \textbf{61.29} & \textbf{48.97} & \textbf{53.84} \\
\bottomrule
\end{tabular}
\vspace{-4mm}
\end{table}

\subsection{What contributes to the understanding improvement?}
\label{sec:exp:attribution}

Table~\ref{tab:attribution} reports the three group means. Adjacent rows are not all single-factor steps: next-scale prediction arrives together with the extra 10B, so that effect is read against the data-matched row.

\noindent\textbf{Data-matched group means.}
\label{sec:exp:data}
Generation-bearing models consume an additional 10B image--text tokens. The data-matched control mixes these 10B tokens into the 55B corpus and trains Qwen-VL on all 65B with image-to-caption supervision only. The three means change by $-0.26$ / $+0.04$ / $-0.04$.

\noindent\textbf{The generation objective matters.}
\label{sec:exp:objective}
Relative to this control, \methodcore{} improves Text / Visual Understanding / Visual Specialty by $+0.99$ / $+1.28$ / $+6.97$. The gain is concentrated on visual specialty. UMM-VQ and Show-o2, retrained under the same budget as paradigm controls, transfer less consistently, especially on text reasoning and the strongest visually grounded tasks.

\noindent\textbf{The target space matters.}
\label{sec:exp:target}
UMM-VQ and \methodcore{} share the extra data and the coarse-to-fine schedule. UMM-VQ predicts discrete DC-HT codes; \methodcore{} predicts continuous SigLIP2 representations. The group means move from 59.64/48.46/51.83 to 59.89/48.55/52.62, and \method{} widens the gap. Perceptual representation targets transfer more directly to understanding. This comparison remains paradigm-level.

\noindent\textbf{Which components carry the gain.}
\label{sec:exp:ablation}
Bidirectional attention alone changes the means only slightly. The jump in these means appears once representation prediction is introduced. Masking further raises visual specialty. Auxiliary reconstruction adds $+1.40$ / $+0.42$ / $+1.22$ and is not the main effect. The no-mask variant cannot initialize the empty 64-scale, so it is scored on understanding only. Per-benchmark numbers are in Appendix~\ref{app:fullablation}.

\subsection{How does generative representation learning change visual evidence use?}
\label{sec:exp:mechanism}

Linear probes are reported for Qwen-VL, the data-matched control, UMM-VQ, Show-o2, and \method{} (Table~\ref{tab:probe}). Qwen-EF is left out because it does not use the SigLIP2 front end. Interventions compare Qwen-VL and \method{}, which share the Qwen3-1.7B LLM, the ViT, and DeepStack injection, but differ in attention, data, and training trajectory, so they compare two complete systems. Every intervention uses the same 300 items and teacher-forced gold-answer scoring. Protocol details are in Appendix~\ref{app:mechprotocol}.

\begin{wraptable}{r}{0.46\textwidth}
\vspace{-3mm}
\centering
\small
\caption{\small Frozen final-layer probes, shared COCO split (Appendix~\ref{app:mechanism:probe}). mIoU: object-mask mean IoU at each patch center. Acc.: same-object grouping accuracy.}
\label{tab:probe}
\begin{tabular}{@{}lcc@{}}
\toprule
Model & mIoU $\uparrow$ & Acc. $\uparrow$ \\
\midrule
Qwen-VL              & 22.97 & 61.28 \\
Qwen-VL data-matched & 23.21 & 61.89 \\
UMM-VQ               & 24.61 & 63.58 \\
Show-o2              & 23.90 & 63.37 \\
\method{}            & \textbf{27.58} & \textbf{68.06} \\
\bottomrule
\end{tabular}
\vspace{-3mm}
\end{wraptable}
\noindent\textbf{Linear probes.}
\label{sec:exp:mechanism:probe}
With the backbone frozen, linear heads on final-layer visual tokens test whether object and spatial structure can be read out without adapting the model (Appendix~\ref{app:mechanism:probe}). Object-mask mIoU measures category-level spatial structure; same-object accuracy measures instance grouping. \method{} reaches 27.58 mIoU and 68.06\% (Table~\ref{tab:probe}). The data-matched Qwen-VL control reaches only 23.21 and 61.89, so the extra image--text data does not explain the increase. UMM-VQ and Show-o2 fall in between: generative training enriches visual states, and predicting perceptual representations preserves more of that structure.

\begin{figure}[t]
\centering
\includegraphics[width=0.31\linewidth]{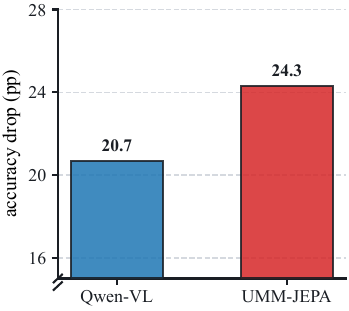}
\hfill
\includegraphics[width=0.31\linewidth]{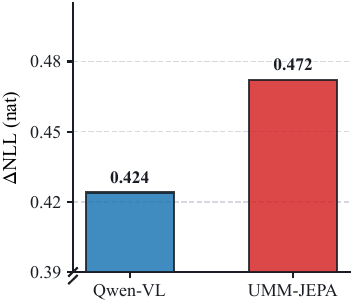}
\hfill
\includegraphics[width=0.31\linewidth]{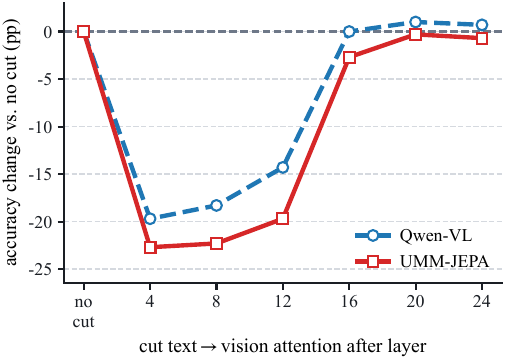}
\caption{\small \textbf{Visual-evidence interventions} on 300 shared items (Appendix~\ref{app:mechprotocol}). Left to right: (a) wrong-image accuracy drop and (b) gold-answer NLL increase, both larger for \method{} under a mismatched image; (c) blocking text-to-vision attention after later layers (uncut accuracy is 65.0 for Qwen-VL and 66.3 for \method{}). Value axes in (a--b) are truncated.}
\label{fig:visual-evidence}
\end{figure}

\noindent\textbf{Wrong image.}
\label{sec:exp:mechanism:wrongimage}
Replacing the image with a mismatched donor from the same pool, at unchanged token count, drops accuracy by 24.3 points for \method{} and 20.7 for Qwen-VL, and raises gold-answer NLL by 0.472 and 0.424, respectively (Figure~\ref{fig:visual-evidence}a,b). The larger drop indicates a stronger dependence on image-specific evidence.

\noindent\textbf{Layer-wise intervention.}
\label{sec:exp:mechanism:intervention}
For every decoder layer with index $>k$ ($k \in \{4, 8, 12, 16, 20, 24\}$), text queries cannot attend to visual keys. Text$\to$text, vision$\to$vision, vision$\to$text, and DeepStack injection into layers 0--2 stay intact, so the probe measures how long text keeps retrieving from visual tokens. Both models depend on that retrieval in early and middle layers. After layer 16, \method{} still loses 2.7 accuracy points, while Qwen-VL returns to about its uncut accuracy (Figure~\ref{fig:visual-evidence}c; Appendix~\ref{app:mechprotocol:intervention}). \method{} keeps retrieving from visual states deeper in the language backbone.

\begin{figure}[t]
\centering
\includegraphics[width=\linewidth]{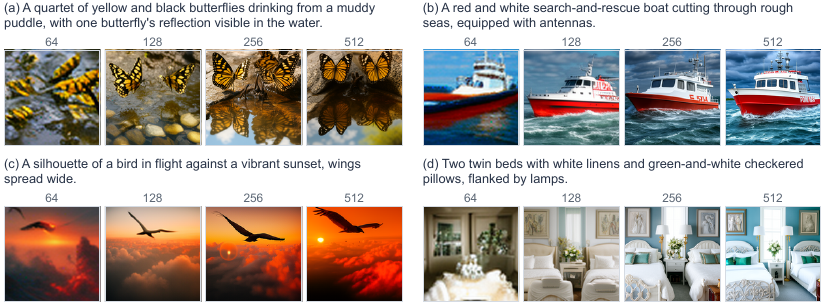}
\caption{\small Progressive generation for four prompts. Each sequence is rendered at scales $64$, $128$, $256$, and $512$. Coarse scales set the layout; later scales add detail. Illustrative only.}
\label{fig:progressive}
\vspace{-3mm}
\end{figure}

Together, the visual states of \method{} retain more linearly decodable object and spatial structure, its answers depend more on the specific image, and it retrieves visual information deeper into the LLM. Patch corruption, token diversity, scale-wise prediction error, and the prediction--understanding correlation are in Appendices~\ref{app:mechprotocol:corruption} and~\ref{app:mechanism}.

\subsection{Image generation results}
\label{sec:exp:generation}

\begin{table}[t]\small
\centering
\begin{minipage}[t]{0.6\textwidth}
\centering
\small
\caption{\small Separate from understanding (Appendix~\ref{app:geneval}). conn.+LoRA+SFT: connector, DiT LoRA, generation SFT. VQ dec.+SFT / latent flow+SFT: native decoder or latent flow, plus SFT. native FM: own renderer. FLUX.2: FLUX.2 [klein]. $\dagger$ Cited.}
\label{tab:gen}
\begin{tabular}{llcc}
\toprule
Model & Pathway & GenEval $\uparrow$ & DPG $\uparrow$ \\
\midrule
Text-only Qwen & conn.+LoRA+SFT & 69.2 & 71.62 \\
Qwen-VL & conn.+LoRA+SFT & 70.5 & 72.41 \\
\method{} & conn.+LoRA+SFT & \textbf{81.2} & \textbf{83.75} \\
UMM-VQ & VQ dec.+SFT & 57.3 & 62.52 \\
Show-o2 & latent flow+SFT & 67.3 & 67.36 \\
FLUX.2 4B$^\dagger$ & native FM & 78.6 & 83.30 \\
BLIP3-o 4B$^\dagger$ & native & 81.0 & 79.36 \\
\bottomrule
\end{tabular}
\end{minipage}
\hfill
\begin{minipage}[t]{0.37\textwidth}
\centering
\small
\caption{\small After the same understanding instruction tuning. Text, VisU, and VisS are means of the 4 text, 7 visual-understanding, and 9 visual-specialty benchmarks in Table~\ref{tab:main}. Per-benchmark results are in Appendix~\ref{app:fullsft}.}
\label{tab:sft}
\begin{tabular}{@{}lccc@{}}
\toprule
Model & Text $\uparrow$ & VisU $\uparrow$ & VisS $\uparrow$ \\
\midrule
Qwen-EF & 60.03 & 35.07 & 20.22 \\
Qwen-VL & 60.63 & 51.99 & 54.74 \\
UMM-VQ & 60.95 & 53.07 & 55.40 \\
Show-o2 & 59.91 & 53.21 & 55.44 \\
\method{} & \textbf{62.76} & \textbf{54.15} & \textbf{58.39} \\
\bottomrule
\end{tabular}
\end{minipage}
\end{table}

Image generation is attached only after representation pretraining (Section~\ref{sec:method:rendering}; Appendix~\ref{app:impl:render}) and is scored separately from understanding. \method{} reaches 81.2 on GenEval and 83.75 on DPG (Table~\ref{tab:gen}). Figure~\ref{fig:progressive} shows the same coarse-to-fine schedule at inference: scales $64$ and $128$ set the layout, and $256$ and $512$ add detail.

Text-only Qwen, Qwen-VL, and \method{} share the same post-hoc pathway, a connector with DiT LoRA and generation SFT. Under that match, \method{} is above Text-only Qwen (69.2/71.62) and Qwen-VL (70.5/72.41). UMM-VQ and Show-o2 are likewise scored after generation SFT, on a VQ decoder and latent flow, at 57.3/62.52 and 67.3/67.36. \method{} stays above both and is competitive with the generation-oriented references.

\subsection{Does the understanding advantage survive instruction tuning?}
\label{sec:exp:sft}

Starting from the Table~\ref{tab:main} checkpoints, every model is fine-tuned on the same Honey-120K understanding data, a 120K subset of Honey-Data-1M \citep{zhang2025bee}, for the same number of updates. \method{} remains strongest, at 62.76 / 54.15 / 58.39 on Text / Visual Understanding / Visual Specialty (Table~\ref{tab:sft}). Qwen-EF stays far behind on both visual groups, consistent with the pretraining comparison. The understanding advantage therefore survives downstream adaptation. Per-benchmark results are in Appendix~\ref{app:fullsft}.

\section{Conclusion}
\label{sec:conclusion}

\method{} learns hierarchical next-scale prediction directly over the continuous representations that understanding uses, with teacher-forced coarse-to-fine training, recursive inference, and pixel rendering kept out of core pretraining. Controlled comparisons show that this improves multimodal understanding beyond an understanding-only baseline, that a data-matched control leaves the three group means nearly unchanged ($-0.26$ / $+0.04$ / $-0.04$), and that perceptual representation targets transfer more consistently than discrete-code or latent-flow objectives; probes and interventions further indicate richer and deeper use of visual evidence. Generation is therefore most useful for understanding when it supervises that shared perceptual space.

\bibliography{iclr2027_conference}

\clearpage
\appendix
\section{Extended related work}
\label{app:related}

This appendix expands the summary of Section~\ref{sec:related} along the four axes that motivate \method{}.

\subsection{Unified understanding and generation}
\label{app:related:unified}

Unified multimodal models adopt markedly different generative parameterizations, including autoregressive discrete visual tokens \citep{chameleon,emu3,janus,song2026broad,zhou2026condition}, next-scale visual autoregression \citep{var}, and latent flow matching or diffusion-based synthesis \citep{showo2,chen2025towards,yang2025dc,yang2026hicogen,wang2026ladr}, together with modality-specialized or partially separated semantic and generative pathways \citep{mot,onecat,bagel,chen2026unify,zhou2026multimodal}. Architectural unification, however, does not determine whether generative training improves the representations used for understanding; we study how the choice of generative target affects that transfer.

\subsection{Generation-to-understanding transfer}
\label{app:related:transfer}

Controlled studies reveal an asymmetric knowledge flow: understanding supervision benefits generation, whereas additional generation training does not consistently improve understanding, with gains depending on the task and reasoning regime \citep{yang2025self,wang2025complexbench,han2026towards,wen2026unig2u,wu2026synergy,zhou2026world}. Auxiliary generation targets such as pixel reconstruction, depth, or segmentation have also been shown to help perception \citep{su2026unimrg}. We ask the complementary and more specific question of whether transfer depends on aligning the \emph{target of generation} with the perceptual representation consumed by understanding; rather than adding several output modalities, we make prediction of that representation the pretraining objective and isolate the choice through data-matched and paradigm controls.

\subsection{Representation-based visual generation}
\label{app:related:repgen}

Prior work places semantically meaningful representations inside the generative pathway, through continuous visual embeddings \citep{emu,emu2}, diffusion in a CLIP feature space \citep{blip3o}, or representation prediction as native intermediate generative states that remain in context to guide pixel diffusion \citep{representationforcing}. We differ in using representation prediction as the object of generative pretraining and in removing image rendering from core pretraining; pixel synthesis is attached only afterward, through a connector and a pretrained diffusion renderer.

\subsection{Predictive representation learning}
\label{app:related:jepa}

Joint-Embedding Predictive Architectures predict target representations from available context instead of reconstructing the observation \citep{ijepa}, showing that prediction in representation space encourages semantic visual representations. This principle motivates our learning target, but our setting is conditional multimodal generation with progressive prediction across a coarse-to-fine hierarchy, and our contribution is to use the principle to reformulate generation as shared representation learning and to study its transfer to understanding.

\section{Compared systems and evaluation}
\label{app:setup}

\noindent\textbf{Models.} Except for the published references marked $\dagger$ in Table~\ref{tab:gen}, every model in our tables is trained by us from the same pretrained Qwen3-1.7B checkpoint. All except the encoder-free reference share the same SigLIP2 visual front end. Generation-bearing controls are trained under the same additional image--text data budget as \method{}, and the Qwen-VL baseline is complemented by a data-matched understanding-only control (Section~\ref{sec:exp:attribution}).
\begin{itemize}
\item \textbf{Qwen-EF}: an encoder-free architecture reference, also initialized from the same pretrained Qwen3-1.7B checkpoint. Only the pretrained ViT is replaced by a patch embedding trained from scratch. It shows the effect of removing the pretrained visual encoder, and is not a controlled comparison of generation objectives.
\item \textbf{Qwen-VL}: our understanding-only run from the same Qwen3-1.7B initialization \citep{qwen3vl}, with causal attention and no additional image--text data.
\item \textbf{UMM-VQ}: keeps the next-scale schedule but predicts discrete DC-HT tokens (4/16/64/256-way classification targets at scales 64/128/256/512) with cross-entropy.
\item \textbf{Show-o2}: retrained with its native unified mechanism \citep{showo2} (latent flow matching over a frozen DC-AE latent space \citep{chen2025dcae}) under the same data budget.
\item \textbf{\methodcore{}}: bidirectional visual attention and masked next-scale representation prediction, without reconstruction.
\item \textbf{\method{}}: \methodcore{} plus auxiliary reconstruction.
\end{itemize}

\noindent\textbf{Data.} All models are trained on the same 10B text and 45B multimodal corpus (55B tokens). Generation-bearing models (UMM-VQ, Show-o2, \methodcore{}, \method{}) additionally mix in 10B image--text tokens (65B total). The data-matched control mixes the same 10B image--text tokens into the 55B corpus and trains Qwen-VL on the combined 65B tokens through image-to-caption understanding supervision only.

\noindent\textbf{Evaluation.} Understanding covers 4 text benchmarks (MMLU \citep{hendrycks2021mmlu}, GSM8K \citep{cobbe2021gsm8k}, MGSM-zh \citep{shi2023mgsm}, MATH-500 \citep{hendrycks2021math,lightman2024verify}), 7 visual-understanding benchmarks (MMMU \citep{yue2024mmmu}, MMMU-Pro \citep{yue2024mmmupro}, MMBench EN/CN \citep{liu2024mmbench}, SEED-Bench \citep{li2023seedbench}, SEED-Bench-2+ \citep{li2024seedbench2plus}, ScienceQA \citep{lu2022scienceqa}), and 9 visual-specialty benchmarks (ChartQA \citep{masry2022chartqa}, OCRBench \citep{liu2023ocrbench}, TextVQA \citep{singh2019textvqa}, DocVQA \citep{mathew2021docvqa}, HallusionBench \citep{guan2024hallusionbench}, VSR \citep{liu2023vsr}, ViewSpatial \citep{li2025viewspatial}, RefCOCO \citep{yu2016refcoco}, CountBench \citep{paiss2023countbench}). \emph{Text}, \emph{Visual Understanding}, and \emph{Visual Specialty} are the means of these 4, 7, and 9 benchmarks, and \emph{Visual} is the mean of all 16. Table~\ref{tab:main} reports every benchmark individually. Shot counts are in Appendix~\ref{app:impl}. Image generation uses GenEval \citep{ghosh2023geneval} and DPG \citep{hu2024ella} (Section~\ref{sec:exp:generation}): GenEval measures compositional fidelity on short prompts, and DPG evaluates entity and relation alignment on longer prompts. The two scores are reported separately and are never averaged with understanding scores. Both are scored locally (Appendix~\ref{app:geneval}).

\section{Implementation details}
\label{app:impl}

\noindent\textbf{Architecture.} Qwen3-1.7B LLM \citep{qwen3} (28 layers, hidden 2048, causal over text) with a SigLIP2 ViT \citep{siglip2} (24 layers, hidden 1024); DeepStack-style injection \citep{qwen3vl} injects ViT features into LLM layers 0/1/2. Every compared run initializes the LLM from the official pretrained Qwen3-1.7B weights. Understanding and generation share all LLM parameters.

\noindent\textbf{Sequence and data.} Packed sequence length 4096 (\method{} with the reconstruction tail: 4224). Data mixture 10B text + 45B multimodal; generation-bearing models and the data-matched control add 10B image--text tokens (10:45:10, 65B total). \method{} stage 2 uses batch size 2000; the Qwen-VL baseline uses a causal LLM with THD packing, batch size 2016, and no additional 10B.

\noindent\textbf{Pretraining stages.} Representation pretraining proceeds in two stages. \emph{Alignment}: the ViT and the LLM are frozen and only the merger is trained on high-quality image--caption pairs for vision--language alignment. \emph{Joint pretraining}: all parameters are unfrozen and trained with cross-entropy and the masked next-scale representation objective of Section~\ref{sec:method:prediction} on 10B text + 45B multimodal tokens, plus 10B image--text tokens for the generation samples; visual patches attend bidirectionally while text remains causal (FlexAttention).

\noindent\textbf{\method{} objective hyperparameters.} Next-scale prediction conditioned on the caption progresses across scales $64, 128, 256,$ and $512$, using frozen SigLIP2 layer-23 features as targets and a cosine loss weighted equally across the four scales ($\lambda_{\mathrm{rep}}=0.55$). For masking, each scale draws its ratio independently from $\mathrm{Uniform}[0.25,0.75]$ and clamps it to $[1,n-1]$ for its token count $n$, and the recipe additionally contains 10\% fully masked samples (matching text-to-image inference, where the $64$-scale is predicted from the caption alone) and 10\% unmasked samples (matching the connector stage, which forbids next-scale masking). At the $128/256$ scales, masked tokens split into about 25\% \emph{inpaint} (ancestors also masked; children predicted purely from holes) and about 75\% \emph{refine} (ancestors visible; finer children predicted on top of coarse content), and masked and visible parents are jointly supervised as $(\mathcal{L}_{\mathrm{masked}} + 0.5\,\mathcal{L}_{\mathrm{visible}})/1.5$ so that the refinement mapping used at inference is trained rather than only the hole-filling mapping. Each scale owns a learnable mask token, while the local projection is shared across scales.

\noindent\textbf{Connector and rendering stages.}
\label{app:impl:render}
The connector is not part of \method{} pretraining; it is trained only after the representation has been learned. Text-only Qwen, Qwen-VL, and \method{} each receive this full recipe, including DiT LoRA and generation SFT, with weights fitted independently. UMM-VQ and Show-o2 keep their native decoders and add generation SFT. The connector architecture and the rendering data are shared across the connector-pathway sources. A $\approx$43.9M connector ($\mathrm{LN} \to \mathrm{Linear}(2048{\to}4096) \to \mathrm{SiLU} \to \mathrm{Linear}(4096{\to}7680) \to \mathrm{LN}$, plus a scale embedding marking which visual span each token comes from, timestep AdaLN, and 10\% CFG dropout to the null condition) translates the 1.7B LLM's 2048-d final-layer hidden states into the 7680-d text-conditioning space of a frozen FLUX.2 [klein] base-4B diffusion transformer \citep{flux2}. The conditioning sequence slices the visual spans by scale: the $64$-scale contributes 4 tokens, the $128$-scale $4{+}16$, and the $256$-scale $4{+}16{+}64$ (the $512$-scale is not encoded).

\noindent\textbf{Stage 1: connector alignment.}
The LLM, SigLIP2, merger, VAE, and DiT are all frozen and only the connector is updated, using the DiT's native rectified-flow objective ($v$-prediction, $x_t = (1{-}t)z + t\epsilon$, logit-normal $t$, per-resolution weights $4{:}4{:}2{:}1$) with gradients flowing through the frozen DiT's activations back to the connector. Training runs for 10{,}000 steps on 8 GPUs with batch size 256, on data randomly sampled from the same 10B image--text generation pool used for representation pretraining. We use Adam ($\beta_1=0.9$, $\beta_2=0.95$, $\epsilon=10^{-8}$, weight decay $0.01$, gradient clipping $1.0$) with a cosine learning-rate schedule from $10^{-4}$ to $10^{-5}$ and 500 warmup steps. Prediction feedback is disabled during early training ($p_{\mathrm{feedback}}=0$) and is then ramped to $p_{\mathrm{feedback}}=0.8$, starting at step 6{,}000 over approximately 1{,}000 steps.

\noindent\textbf{Stage 2: renderer adaptation with prediction feedback.}
The multimodal backbone, the VAE, the DiT main weights, the text-modulation modules, the image branch, and the single-stream blocks all remain frozen. We optimize only the connector together with lightweight LoRA adapters inserted into the text-conditioning modules of the diffusion transformer: rank $r=16$ with $\alpha=16$ (so the scaling factor is $1$) on \texttt{txt\_in} and on five double blocks, targeting \texttt{txt\_attn.qkv}, \texttt{txt\_attn.proj}, \texttt{txt\_mlp.0}, and \texttt{txt\_mlp.2}. The adapters contain $4{,}349{,}952\approx4.3$M parameters and, together with the $\approx43.9$M connector, give $\approx48.2$M trainable parameters. Every step is trained with prediction feedback ($p_{\mathrm{feedback}}=1.0$) using a two-pass procedure. In the first pass, the frozen representation pathway predicts the $64/128/256$ scales in a single denoising step with $t\sim\mathcal{U}(0.15,0.45)$ and without gradient computation; the resulting one-step $x_0$ estimates are spliced back into the conditioning sequence as visual context. The second pass recomputes the forward pass from this predicted context and evaluates the diffusion velocity objective over the $64/128/256/512$ scales with per-resolution weights $4{:}4{:}2{:}1$, and only this second-pass loss is back-propagated. Training runs for 10{,}000 steps on 8 GPUs with batch size 256 (no gradient accumulation) and CFG dropout $0.1$, using the same Adam configuration, gradient clipping, and data source as above, with a cosine learning-rate schedule from $10^{-5}$ to $10^{-6}$ and 200 warmup steps.

\noindent\textbf{Stage 3: generation instruction tuning.}
Starting from the adapted connector and diffusion adapters, we continue training the connector and the DiT LoRA adapters and additionally introduce LoRA adapters on the Qwen attention $q/k/v/o$ projections. The Text-only Qwen, Qwen-VL, and \method{} entries in Table~\ref{tab:gen} are taken from the checkpoint after this stage (6{,}000 steps). UMM-VQ and Show-o2 are scored from their own decoders after generation SFT. FLUX.2 and BLIP3-o are published references.

\noindent\textbf{Evaluation protocol.} Shots per benchmark: MMLU 0, GSM8K 0, MGSM-zh 8, MATH-500 0; MMMU 3, MMMU-Pro 3, MMBench EN 3, MMBench CN 0, SEED-Bench 0, SEED-Bench-2+ 3, ScienceQA 0; ChartQA 0, OCRBench 0, TextVQA 0, DocVQA 0, HallusionBench 0 (aAcc), VSR 0, ViewSpatial 3, RefCOCO 0, CountBench 0. All columns of Table~\ref{tab:main} use the same protocol per benchmark.

\noindent\textbf{Reconstruction tail in \method{}.}
\label{app:impl:rec}
The tail is appended only to understanding samples; generation samples are unchanged, and the packed sequence length grows from 4096 to 4224 ($+128$ tokens). The original packed understanding row is kept intact and a reconstruction segment is appended to its end: a random instruction, a \texttt{<gen>} marker, the $64$, $128$, and $256$ visual spans, and a closing \texttt{</gen>} marker. The target image is the last image of that packed row, and the entire reconstruction segment carries $\mathrm{CE\ label}=-100$, so it does not contribute to the language-modeling loss.

\noindent\textbf{Visual isolation.} FlexAttention isolates the original image tokens preceding \texttt{<gen>} from the reconstruction segment, so the reconstruction cannot directly attend back to the target image and must rebuild that image's visual representations from the textual context of the understanding sample alone.

\noindent\textbf{Targets, scales, and loss.} The teacher is frozen SigLIP2 layer 23, i.e., the last layer of the 24-layer tower, and its features are LayerNorm-ed before serving as targets; the student prediction is read from the LLM decoder hidden states. The four scales ($64, 128, 256,$ and $512$) are equally weighted ($1{:}1{:}1{:}1$) and normalized, following the same coarse-to-fine progression initialized from \texttt{<gen>} as in pretraining (where the $512$ scale serves only as a prediction target rather than an input span). Visible parents receive weight $0.5$ relative to masked parents. The loss is the cosine discrepancy $\mathcal{L}_{\mathrm{rec}}=1-\cos(\hat{\mathbf{z}},\mathbf{z})$, not MSE or L1, with weight $\lambda_{\mathrm{rec}}=0.55$.

\noindent\textbf{Mask recipe.} 10\% of reconstruction samples are fully masked and 10\% are unmasked; the remaining samples use connected-block masking with the ratio drawn from $[0.25,0.75]$, which yields the approximately 25\% inpaint setting of the main objective. No context noise, blur, or cutout is applied, and the understanding-side representation loss is disabled outside the reconstruction tail.

\noindent\textbf{Training schedule: representation pretraining.}
\label{app:impl:schedule:pretrain}
The reported run is the stage-2 joint representation pretraining, initialized from the SigLIP2 v1 stage-1 alignment checkpoint; no separate alignment stage is run for this model. The final checkpoint is at iteration 7{,}694. The learning rate follows a cosine schedule with 125 warmup steps, decaying from $6\times10^{-5}$ to $6\times10^{-6}$ for the LLM and from $3\times10^{-5}$ to $3\times10^{-6}$ for the ViT and merger, using the Adam configuration given for the rendering stages. Batch size is fixed at 2000 on 8 GPUs with no batch-size ramp. Sequences are packed to $4224 = 4096 + 128$, where the 128 tokens are the reconstruction-tail headroom. The token budget is 45B multimodal, 10B text, and 10B generation image--text tokens.

\noindent\textbf{Training schedule: rendering stages.} The frozen-renderer connector stage and the renderer-adaptation stage use the configurations given above (10{,}000 steps each; 8 GPUs, batch size 256, $10^{-4}\to10^{-5}$ cosine with 500 warmup and feedback $0\to0.8$ from step 6{,}000 for the connector; 8 GPUs, batch size 256, $10^{-5}\to10^{-6}$ cosine with 200 warmup and $p_{\mathrm{feedback}}=1.0$ with a two-pass step for renderer adaptation). Both stages sample their images randomly from the same 10B image--text generation pool.

\noindent\textbf{Training schedule: understanding instruction tuning.} Understanding instruction tuning uses the Honey-120K subset of Honey-Data-1M \citep{zhang2025bee} with 45{,}765 packed rows for 2 epochs. The learning rate follows a cosine schedule with 35 warmup steps, from $10^{-5}$ to $10^{-6}$ for the LLM and from $5\times10^{-6}$ to $5\times10^{-7}$ for the ViT. Batch size is 256 on 8 GPUs with a packed sequence length of 4096.

\section{Theoretical Analysis}
\label{app:theory}

\subsection{Assumptions and Notation}

We consider an image $\mathbf{x}=\mathcal{R}(\mathbf{s},\mathbf{n})$ generated from a task-relevant latent scene state $\mathbf{s}$ and rendering-specific nuisance variables $\mathbf{n}$, a perceptual encoder $\mathbf{z}=E(\mathbf{x})$, and an understanding mapping $\mathbf{y}=h(\mathbf{z})$. Vectors are compared with the Euclidean norm, $\sigma(\cdot)$ denotes the $\sigma$-algebra generated by a random variable, and risk is measured under squared loss. The analysis assumes:
\begin{itemize}
    \item \emph{Locally regular understanding mapping.} On the region of representation space visited during training, $h$ is $L$-Lipschitz.
    \item \emph{Squared prediction loss.} Prediction error is measured by squared Euclidean distance in representation space.
    \item \emph{Nested conditioning.} For each scale, the coarse conditioning variable $\mathcal{C}_k=(C,\mathbf{Z}^{64},\ldots,\mathbf{Z}^{k})$ generates at least the information of the text condition $C$, i.e., $\sigma(C)\subseteq\sigma(\mathcal{C}_k)$.
    \item \emph{Masked target exclusion.} When the states in $M$ are masked, they are removed from the model's observation, and the observed variables are $C$ and $\mathbf{Z}_{\bar M}$.
\end{itemize}
Nothing below relies on a parametric form for $E$, $h$, or the data distribution.

\subsection{Proof of Proposition~\ref{prop:target}}

\begin{proof}
By (A1), $h$ is $L$-Lipschitz on the relevant region, so for any predicted state $\hat{\mathbf{z}}$ and target $\mathbf{z}$ we have $\|h(\hat{\mathbf{z}})-h(\mathbf{z})\| \le L\|\hat{\mathbf{z}}-\mathbf{z}\|$. Squaring and taking expectations,
\begin{align}
    \mathbb{E}\|h(\hat{\mathbf{z}})-h(\mathbf{z})\|^2 \le L^2\,\mathbb{E}\|\hat{\mathbf{z}}-\mathbf{z}\|^2,
\end{align}
so the deviation of the downstream prediction from its target is controlled by the representation-space prediction error itself. The constant is tight: for a linear mapping $h(\mathbf{z})=W\mathbf{z}$ with $\|W\|_2=L$, equality holds whenever $\hat{\mathbf{z}}-\mathbf{z}$ aligns with the leading singular direction of $W$. The statement bounds the sensitivity of the understanding output to errors in the state that understanding consumes; it does not compare this bound with the sensitivity of any other prediction target.
\end{proof}

\subsection{Proof of Lemma~\ref{lem:coarse}}

\begin{proof}
Under squared loss, the minimum achievable risk of predicting $\mathbf{Z}^{k+1}$ from a conditioning variable $U$ is the expected conditional variance,
\begin{align}
    \mathcal{R}^*(\mathbf{Z}^{k+1}\mid U)
    =
    \mathbb{E}\big[\|\mathbf{Z}^{k+1}-\mathbb{E}[\mathbf{Z}^{k+1}\mid U]\|^2\big]
    =
    \mathbb{E}\big[\operatorname{Var}(\mathbf{Z}^{k+1}\mid U)\big],
\end{align}
since the conditional mean is the Bayes-optimal predictor. By (A3), $\sigma(C)\subseteq\sigma(\mathcal{C}_k)$, so $\mathcal{C}_k$ is an enrichment of $C$. The law of total variance gives, pointwise,
\begin{align}
    \operatorname{Var}(\mathbf{Z}^{k+1}\mid C)
    =
    \operatorname{Var}\big(\mathbb{E}[\mathbf{Z}^{k+1}\mid\mathcal{C}_k]\mid C\big)
    +
    \mathbb{E}\big[\operatorname{Var}(\mathbf{Z}^{k+1}\mid\mathcal{C}_k)\mid C\big]
    \;\ge\;
    \mathbb{E}\big[\operatorname{Var}(\mathbf{Z}^{k+1}\mid\mathcal{C}_k)\mid C\big].
\end{align}
Taking expectations and substituting the risk identity yields
\begin{align}
    \mathcal{R}^*(\mathbf{Z}^{k+1}\mid C)
    \;\ge\;
    \mathbb{E}\big[\operatorname{Var}(\mathbf{Z}^{k+1}\mid\mathcal{C}_k)\big]
    =
    \mathcal{R}^*(\mathbf{Z}^{k+1}\mid\mathcal{C}_k).
\end{align}
Conditioning on additional coarse visual states can therefore only reduce the Bayes-optimal fine-scale risk, and it does so strictly whenever those states carry information about the fine scale beyond the text condition. The lemma constrains the value of the conditioning variables, not which scales or how many of them are used; the scale schedule is an implementation choice.
\end{proof}

\subsection{Proof of Proposition~\ref{prop:mask}}

\begin{proof}
Under squared loss and (A4), the observed variables for a masked prediction are $C$ and $\mathbf{Z}_{\bar M}$, and the risk of a predictor $f$ is
\begin{align}
    \mathbb{E}\big[\|f(C,\mathbf{Z}_{\bar M})-\mathbf{Z}_M\|^2\big],
\end{align}
which is finite and convex in $f$. The minimizer is therefore obtained pointwise, since the choice of $f$ at one value of $(C,\mathbf{Z}_{\bar M})$ does not affect the integrand elsewhere,
\begin{align}
    f^*(C,\mathbf{Z}_{\bar M})
    =
    \arg\min_{v}\ \mathbb{E}\big[\|\mathbf{Z}_M-v\|^2 \mid C,\mathbf{Z}_{\bar M}\big]
    =
    \mathbb{E}[\mathbf{Z}_M\mid C,\mathbf{Z}_{\bar M}].
\end{align}
Because $\mathbf{Z}_M$ is excluded from the observation, the optimum must be formed from the caption and the visible states alone, i.e., from textual, spatial, and cross-scale context. If $M$ is the full state set, (A4) leaves $C$ as the only observation and the predictor degenerates to $f^*(C)=\mathbb{E}[\mathbf{Z}\mid C]$, the text-only case discussed after Proposition~\ref{prop:mask} in the main text.
\end{proof}

\subsection{Scope of the Analysis}

The analysis motivates the choice of prediction target, progressive conditioning, and masking under simplified prediction assumptions. It does not claim that the particular scale schedule or masking ratio is theoretically optimal; these choices are evaluated empirically.

\section{Full per-benchmark results}
\label{app:fullresults}

\subsection{Encoder-free reference (architecture sanity check)}
\label{app:arch}

Table~\ref{tab:arch} provides the full pretraining and instruction-tuning results of the encoder-free architecture reference. Qwen-EF is initialized from the same pretrained Qwen3-1.7B checkpoint as the other systems and replaces only the SigLIP2 front end with a patch embedding trained from scratch. Table~\ref{tab:arch} reports its per-benchmark results before and after understanding instruction tuning. It collapses on the visually grounded specialty benchmarks (for example 2.70 vs.\ 48.30 on OCRBench at pretraining), which is why a pretrained visual front end is treated as a prerequisite at this scale rather than as a variable under study.

\begin{table}[t]\small
\centering
\caption{\small Encoder-free architecture reference (Qwen-EF), full per-benchmark results. Qwen-EF is included as an architecture reference in Table~\ref{tab:main}; the table here reports its complete before/after instruction-tuning breakdown.}
\label{tab:arch}
\begin{tabular}{llcc}
\toprule
 & Benchmark & Qwen-EF (pretraining) & Qwen-EF (instr.\ tuned) \\
\midrule
\multirow{4}{*}{\shortstack[l]{Text}}
 & MMLU                & 49.54 & 50.18 \\
 & GSM8K               & 76.12 & 77.33 \\
 & MGSM-zh             & 59.20 & 62.00 \\
 & MATH-500            & 53.20 & 50.60 \\
\midrule
\multirow{7}{*}{\shortstack[l]{Visual\\understanding}}
 & MMMU                & 43.72 & 39.67 \\
 & MMMU-Pro            & 20.17 & 20.29 \\
 & MMBench EN          & 14.86 & 16.41 \\
 & MMBench CN          & 12.00 & 15.25 \\
 & SEED-Bench          & 40.13 & 40.61 \\
 & SEED-Bench-2+       & 42.69 & 42.51 \\
 & ScienceQA           & 69.21 & 70.75 \\
\midrule
\multirow{9}{*}{\shortstack[l]{Visual\\specialty}}
 & ChartQA             & 9.32  & 12.44 \\
 & OCRBench            & 2.70  & 2.60 \\
 & TextVQA             & 5.78  & 7.29 \\
 & DocVQA              & 6.10  & 7.08 \\
 & HallusionBench      & 54.26 & 57.83 \\
 & VSR                 & 51.88 & 50.08 \\
 & ViewSpatial         & 28.17 & 29.24 \\
 & RefCOCO             & 7.55  & 6.97 \\
 & CountBench          & 11.50 & 8.42 \\
\bottomrule
\end{tabular}
\end{table}

\subsection{Understanding instruction tuning per benchmark}
\label{app:fullsft}

Table~\ref{tab:fullsft} reports the per-benchmark results behind the group means of Table~\ref{tab:sft}. Several of the largest pretraining gaps on visual specialty narrow after this tuning: ChartQA moves from $+15.44$ to $+0.04$, and RefCOCO from $+13.91$ to $+2.41$. OCRBench, TextVQA, DocVQA, and CountBench keep a clearer margin, and the group means in Table~\ref{tab:sft} remain the highest.

\begin{table}[t]\small
\centering
\caption{\small Per-benchmark results after understanding instruction tuning.}
\label{tab:fullsft}
\begin{tabular}{llcccc}
\toprule
 & Benchmark & Qwen-VL & UMM-VQ & Show-o2 & \method{} \\
\midrule
\multirow{4}{*}{\shortstack[l]{Text}}
 & MMLU                & 51.91 & 49.59 & 50.47 & \textbf{52.76} \\
 & GSM8K               & 76.42 & 76.19 & 76.57 & \textbf{77.48} \\
 & MGSM-zh             & 63.60 & 62.40 & 62.40 & \textbf{64.00} \\
 & MATH-500            & 50.60 & 55.60 & 50.20 & \textbf{56.80} \\
\midrule
\multirow{7}{*}{\shortstack[l]{Visual\\understanding}}
 & MMMU                & 40.50 & 43.17 & 43.00 & \textbf{44.94} \\
 & MMMU-Pro            & 22.43 & 22.77 & 23.06 & \textbf{23.53} \\
 & MMBench EN          & 53.56 & 54.02 & 55.65 & \textbf{56.58} \\
 & MMBench CN          & 53.02 & 53.41 & 53.95 & \textbf{54.41} \\
 & SEED-Bench          & 65.77 & 66.07 & 65.73 & \textbf{66.29} \\
 & SEED-Bench-2+       & 54.72 & 54.28 & 53.97 & \textbf{54.90} \\
 & ScienceQA           & 73.92 & 77.74 & 77.14 & \textbf{78.38} \\
\midrule
\multirow{9}{*}{\shortstack[l]{Visual\\specialty}}
 & ChartQA             & 62.76 & 61.60 & 62.76 & \textbf{62.80} \\
 & OCRBench            & 57.80 & 58.30 & 58.00 & \textbf{62.40} \\
 & TextVQA             & 61.16 & 61.89 & 61.66 & \textbf{65.41} \\
 & DocVQA              & 53.48 & 52.39 & 52.39 & \textbf{60.40} \\
 & HallusionBench      & 61.09 & 61.30 & 61.72 & \textbf{62.25} \\
 & VSR                 & 48.53 & 48.53 & 48.53 & \textbf{51.39} \\
 & ViewSpatial         & 33.05 & 33.14 & 34.40 & \textbf{35.08} \\
 & RefCOCO             & 46.17 & 46.33 & 43.12 & \textbf{48.58} \\
 & CountBench          & 68.58 & 75.15 & 76.39 & \textbf{77.21} \\
\bottomrule
\end{tabular}
\end{table}

\section{Full component ablations}
\label{app:fullablation}

Table~\ref{tab:fullablation} reports the per-benchmark results behind the group means of the component rows of Table~\ref{tab:attribution}. Bold marks the best among the five columns. Auxiliary reconstruction is uneven across benchmarks: relative to the w/o Recon column, \method{} is lower on MMBench EN, ScienceQA, ChartQA, DocVQA, and VSR. The group-mean increase comes mainly from text and from RefCOCO ($39.80 \rightarrow 46.53$).

\begin{table}[t]\small
\centering
\caption{\small Per-benchmark component-ablation results (pretraining stage).}
\label{tab:fullablation}
\begin{tabular}{llccccc}
\toprule
 & Benchmark & Qwen-VL & +Bi-Attn & w/o Mask & w/o Recon & \method{} \\
\midrule
\multirow{4}{*}{\shortstack[l]{Text}}
 & MMLU                & 51.23 & 51.35 & 51.36 & 51.40 & \textbf{51.57} \\
 & GSM8K               & 76.42 & 75.66 & 75.82 & 75.97 & \textbf{76.80} \\
 & MGSM-zh             & 59.20 & 60.00 & 60.40 & 60.00 & \textbf{61.60} \\
 & MATH-500            & 49.80 & 51.60 & 52.00 & 52.20 & \textbf{55.20} \\
\midrule
\multirow{7}{*}{\shortstack[l]{Visual\\understanding}}
 & MMMU                & 41.61 & 43.00 & 41.89 & 42.11 & \textbf{45.11} \\
 & MMMU-Pro            & 20.64 & 21.18 & 21.16 & 21.39 & \textbf{21.50} \\
 & MMBench EN          & 47.76 & 45.67 & 48.03 & \textbf{48.07} & 47.76 \\
 & MMBench CN          & 33.98 & 37.31 & 37.28 & 37.31 & \textbf{37.54} \\
 & SEED-Bench          & 60.37 & 61.64 & 62.25 & 62.26 & \textbf{62.42} \\
 & SEED-Bench-2+       & 51.43 & 51.60 & 52.61 & 52.17 & \textbf{52.88} \\
 & ScienceQA           & 74.81 & 74.17 & 75.71 & \textbf{76.55} & 75.61 \\
\midrule
\multirow{9}{*}{\shortstack[l]{Visual\\specialty}}
 & ChartQA             & 41.20 & 41.60 & 56.12 & \textbf{58.00} & 56.64 \\
 & OCRBench            & 48.30 & 48.50 & 55.00 & 53.30 & \textbf{55.90} \\
 & TextVQA             & 48.83 & 48.72 & 56.92 & 57.23 & \textbf{58.12} \\
 & DocVQA              & 31.36 & 31.67 & 41.67 & \textbf{43.71} & 43.09 \\
 & HallusionBench      & 60.25 & 58.99 & 58.99 & 61.41 & \textbf{62.04} \\
 & VSR                 & 49.02 & 46.97 & 52.54 & \textbf{56.06} & 55.56 \\
 & ViewSpatial         & 31.48 & 32.49 & 33.30 & 32.04 & \textbf{34.21} \\
 & RefCOCO             & 32.62 & 32.50 & 36.90 & 39.80 & \textbf{46.53} \\
 & CountBench          & 68.17 & 70.84 & 71.46 & 72.07 & \textbf{72.48} \\
\bottomrule
\end{tabular}
\end{table}

\FloatBarrier
\section{Visual-evidence intervention protocol}
\label{app:mechprotocol}

Wrong-image replacement, patch corruption, and the layer-wise attention cut in this section share 300 items with the token-diversity, scale-error, and prediction--understanding analyses in Appendix~\ref{app:mechanism}. The 300 items are drawn once and fixed across probes and arms. Scoring is teacher-forced: each item is scored twice per condition, and the metric is the full-vocabulary probability of the gold answer letter at the answer position (accuracy by argmax; NLL as the mean token-level negative log-likelihood of the gold letter).

\subsection{Wrong image}
\label{app:mechprotocol:wrongimage}

The input image is replaced by a mismatched donor drawn from the same item pool; the donor is resized (bicubic) to the target's pixel size so the visual-token count is unchanged. Prompt, few-shot prefix, and gold answer are held fixed. Acc drop $=$ Acc(correct) $-$ Acc(wrong); NLL increase $=$ NLL(wrong) $-$ NLL(correct). Figure~\ref{fig:visual-evidence}(a,b) reports the results.

\subsection{Corruption}
\label{app:mechprotocol:corruption}

Before the SigLIP2 front-end, $k = \mathrm{round}(N_{\text{patch}} \cdot r)$ patch positions are zeroed with $r \in \{10\%, 30\%, 50\%\}$, plus a clean baseline; grid size and token count are unchanged. The random mask is identical between the two arms. This is random patch masking, not the training-time NSP cascade mask.

\begin{figure}[t]
\centering
\includegraphics[width=0.54\linewidth]{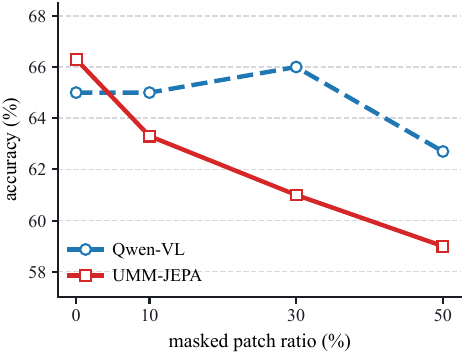}
\caption{\small Corruption probe on the same 300 items: accuracy after zeroing $r$ patch positions before the SigLIP2 front end ($r \in \{0,10,30,50\}\%$), with grid size, token count, and the random mask held identical across the two arms. Qwen-VL is essentially flat ($-2.3$ points at 50\%), whereas \method{} falls by 7.3 points.}
\label{fig:corruption}
\end{figure}

Figure~\ref{fig:corruption} reports the accuracy under patch corruption. \method{} is \emph{more sensitive} to the removal of visual evidence: at 50\% corruption its accuracy falls by 7.3 points, whereas Qwen-VL falls by 2.3. This is not a robustness result and we do not present it as one. It is consistent with the wrong-image probe in indicating stronger reliance on the actual visual content, but it is equally compatible with lower robustness to missing pixels, so it is reported here rather than in the main text.

\subsection{Layer-wise intervention}
\label{app:mechprotocol:intervention}

At every LLM decoder layer with 1-based index $> k$ ($k \in \{4,8,12,16,20,24\}$; 28 layers total), attention scores from text queries to visual keys are set to $-\infty$; text$\to$text, vision$\to$vision (including \method{}'s within-span bidirectional attention in uncut layers), vision$\to$text, causality, and padding are preserved. Visual keys are image-token positions; vision start/end markers are treated as text. DeepStack residuals injecting ViT features into LLM layers 0--2 are unaffected at every $k$, so the probe measures late direct retrieval only. Drop $=$ uncut accuracy $-$ cut accuracy. Table~\ref{tab:layerwise} reports the per-$k$ drop, also shown as Figure~\ref{fig:visual-evidence}(c).

\begin{table}[t]\small
\centering
\caption{\small Layer-wise intervention: accuracy drop (percentage points) after blocking text$\to$vision attention at layers $> k$, on the same 300 items (uncut accuracy 65.0 for Qwen-VL, 66.3 for \method{}).}
\label{tab:layerwise}
\begin{tabular}{lcccccc}
\toprule
Model & $k{=}4$ & $k{=}8$ & $k{=}12$ & $k{=}16$ & $k{=}20$ & $k{=}24$ \\
\midrule
Qwen-VL   & 19.7 & 18.3 & 14.3 & 0.0 & $-1.0$ & $-0.7$ \\
\method{} & 22.7 & 22.3 & 19.7 & 2.7 & 0.3 & 0.7 \\
\bottomrule
\end{tabular}
\end{table}

\FloatBarrier
\section{Additional representation analyses}
\label{app:mechanism}

\subsection{Linear probe protocol}
\label{app:mechanism:probe}

The LLM and the visual front end are frozen and only the linear heads are trained. We use the final-layer visual-token hidden states (not the DeepStack-injected early layers). Training images are 500 images stratified over the 80 COCO categories \citep{lin2014coco} from \texttt{train2017} and validation images are 1000 images stratified the same way from \texttt{val2017} (seed 42), shared by all models. The object-mask probe assigns each visual patch the COCO semantic category of the instance whose mask covers the patch center, and background (class 0) when the center falls outside every instance, giving $80+1=81$ classes; when instances overlap, the smaller-area instance takes precedence. It is scored by mIoU. The same-object probe instead samples visual-token pairs within an image (up to 16 positive and 16 negative pairs per image) and predicts whether both patches belong to the same object, using the per-image instance identity as the label; instance identity does not enter the mIoU target. It is scored by accuracy. The object-mask probe therefore evaluates category-level spatial structure, whereas the same-object probe evaluates instance-level grouping. The head is trained with AdamW at learning rate $1\times10^{-3}$ for 20 epochs for every model. Table~\ref{tab:probe} reports the results.

\subsection{Late-layer token diversity}
\label{app:mechanism:diversity}

\begin{figure}[t]
\centering
\includegraphics[width=0.65\linewidth]{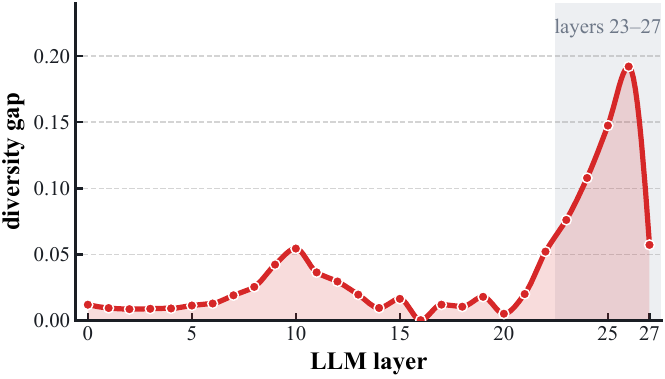}
\caption{\small Per-layer visual-token diversity gap (\method{} minus Qwen-VL). Diversity at each layer is $1 -$ the mean pairwise cosine among that layer's visual tokens, and each point is the mean gap over the 300 shared items. The shaded band marks layers 23--27, whose means are 0.486 for \method{} and 0.370 for Qwen-VL.}
\label{fig:diversity}
\end{figure}

For each layer, token diversity $= 1 - $ mean pairwise cosine among visual tokens, computed in the same forward pass for both arms. \method{} reaches 0.486 versus 0.370 for Qwen-VL (Figure~\ref{fig:diversity}) when averaging layers 23--27, i.e., late-layer visual states remain more differentiated. This is a descriptive pairwise statistic and does not by itself demonstrate richer semantic content.

\subsection{Scale-wise prediction error}
\label{app:mechanism:scale}

\begin{figure}[t]
\centering
\includegraphics[width=0.60\linewidth]{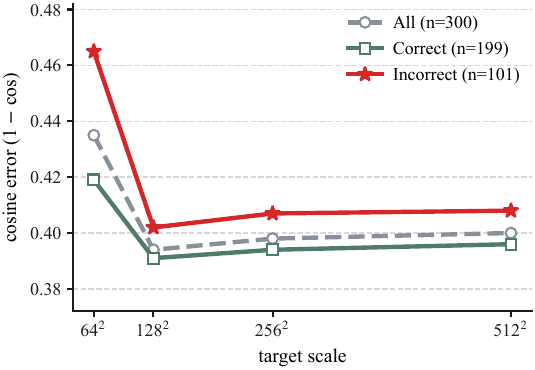}
\caption{\small Scale-wise next-scale prediction error of \method{} ($1-$cosine) on the 300 shared items, split by whether the item was answered correctly: mean 0.435 / 0.394 / 0.398 / 0.400, correct ($n{=}199$) 0.419 / 0.391 / 0.394 / 0.396, incorrect ($n{=}101$) 0.465 / 0.402 / 0.407 / 0.408 at scales 64 / 128 / 256 / 512.}
\label{fig:scale}
\end{figure}

For \method{} only, each item additionally runs an all-mask NSP forward in which the item's own image is the generation target and the conditioning text is the item's question. Per-scale error is $1-$cosine: predictions are projected to the teacher dimension, both prediction and frozen SigLIP2 target are length-normalized, and the inner product is averaged over tokens (training weight 0.55 not applied). Figure~\ref{fig:scale} reports the per-scale errors. The error drops from 0.435 at scale 64 to 0.394 at scale 128 and then plateaus near 0.40, and correct and incorrect items separate mainly at the coarsest scale (0.419 vs.\ 0.465 at 64, and about 0.01 thereafter). In other words, the coarse scale already carries much of the understanding-relevant signal.

\subsection{Prediction--understanding correlation}
\label{app:mechanism:corr}

\begin{figure}[t]
\centering
\includegraphics[width=0.42\linewidth]{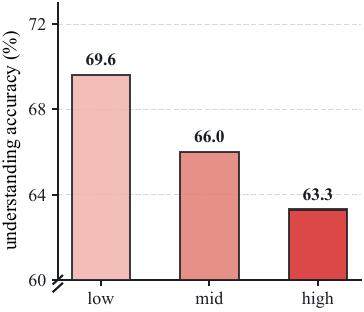}
\caption{\small Understanding accuracy (\%) by tercile of the 512-scale prediction error (300 items), split at the $1/3$ and $2/3$ quantiles. The value axis is truncated and the break is marked.}
\label{fig:corr}
\end{figure}

Using the 512-scale errors of the same 300 items, we split the items into low/mid/high terciles by the $1/3$ and $2/3$ quantiles and report the understanding accuracy of each tercile (Figure~\ref{fig:corr}). Accuracy decreases from 69.6 to 66.0 to 63.3 with increasing prediction error, i.e., better next-scale prediction is associated with higher understanding accuracy, but the effect is modest and correlational: these analyses do not establish that better prediction causes better understanding.

\section{Generation evaluation protocol}
\label{app:geneval}

Generation is evaluated with two metrics, both scored locally with no paid APIs. \textbf{GenEval} \citep{ghosh2023geneval}: the official 553 prompts covering objects, counts, colors, and positions, scored by a local Mask2Former \citep{cheng2022mask2former} (raw 0--1 score scaled by 100). \textbf{DPG} \citep{hu2024ella}: 1065 long prompts probing entities and relations, scored by a local mPLUG-large \citep{li2022mplug} (natively 0--100). The two metrics are reported separately and are never averaged with understanding scores.